# Optical Font Recognition in Smartphone-Captured Images, and its Applicability for ID Forgery Detection

Yulia S. Chernyshova[1,2], Mikhail A. Aliev[1,2], Ekaterina S. Gushchanskaia[3] and Alexander V. Sheshkus[1,2]

[1] Federal Research Center "Computer Science and Control" of Russian Academy of Sciences
Institute for Systems Analysis, Moscow, Russia
[2] Smart Engines, Moscow, Russia
[3] Boston University, Boston, USA

## ABSTRACT

In this paper, we consider the problem of detecting counterfeit identity documents in images captured with smartphones. As the number of documents contain special fonts, we study the applicability of convolutional neural networks (CNNs) for detection of the conformance of the fonts used with the ones, corresponding to the government standards. Here, we use multi-task learning to differentiate samples by both fonts and characters and compare the resulting classifier with its analogue trained for binary font classification. We train neural networks for authenticity estimation of the fonts used in machine-readable zones and ID numbers of the Russian national passport and test them on samples of individual characters acquired from 3238 images of the Russian national passport. Our results show that the usage of multi-task learning increases sensitivity and specificity of the classifier. Moreover, the resulting CNNs demonstrate high generalization ability as they correctly classify fonts which were not present in the training set. We conclude that the proposed method is sufficient for authentication of the fonts and can be used as a part of the forgery detection system for images acquired with a smartphone camera.

## 1. INTRODUCTION

Documents containing personal data are required for many services. The most obvious and potentially dangerous ID forgery occurs at the border control, where manual document inspection becomes almost impossible with constantly increasing passenger traffic [1]. When it comes to financial services [2], banks, insurance and telecommunication companies suffer huge losses. According to the Russian National Credit bureau, annual losses in the credit segment alone exceed 150 billion of Russian rubles [3]. In spite of all protection measures, the number of forged documents increases every year [4].

In the meantime, more and more companies develop services that require ID images captured with smartphones and other mobile devices [5]. Thus, the problem of automatic authentication of documents in photos in the visible spectrum becomes more and more relevant.

Most common type of document forgery (identity documents included), especially while using digital ways of data transmission, is text data falsification [6]. For a number of identity documents, particular fonts are specified by the government decrees and standards, for example, OCR-B font used within machine-readable zones [7]. Since a number of the special fonts are not publically available or officially classified, fraudsters use fonts that are similar but not identical to the specified ones. For instance, Fig.1 (a) demonstrates a line from a machine-readable zone (MRZ) of a counterfeit document printed with a forged font and Fig.1 (b) shows its version meeting the standards [7].

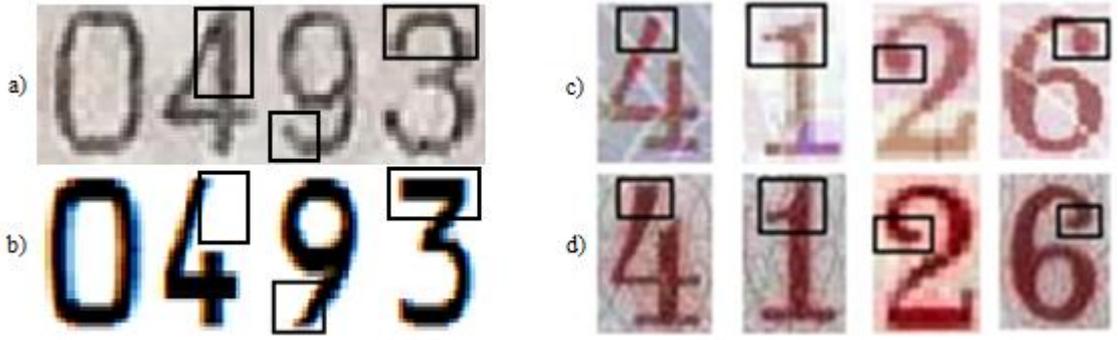

Figure 1. a) Character examples from a forged MRZ; b) genuine versions of MRZ symbols; c) character examples from a forged Russian passport ID number; d) genuine version of symbols from a Russian passport ID number

The aim of our study is to develop a neural network method to determine the conformance of a font in the document with a standard one using images of separate symbols captured with a smartphone camera. It is essential that the method should demonstrate adequate results on fonts which are used as the replacement for the one specified in the standards but are not presented in the training set.

This paper is structured as follows: in Section 2 we provide an overview of forgery detection methods for documents in images and the methods of optical font recognition. In Section 3 we formulate a problem statement and describe its special traits. In Section 4 we describe train and test data and discuss the experiments.

## 2. RELATED WORK

### 2.1 Methods of detecting forgery of the documents based on images

In the previous study [1], the correspondence between data in the main part of the document and data in the machine-readable zone is analyzed, and also cross-verification is performed for images of the same class. Other works [8,9] propose detection methods for "copy-paste" type forgery, i.e. the authors use pixel analysis to detect the similarity among symbols that were copied and repositioned within the same document. Additionally, some authors suggest using the typographical features of fonts[10]. Based on their conclusions [10], other researchers propose to use Conditional Random Fields to get the estimation of text font by its typographical features and to assess its correspondence with other text within the document. However, previously applied methods cannot be fully embedded in a system for images acquired with small-scale digital cameras in uncontrolled conditions [12], as those approaches assume that images are captured with an even distribution of light [1], and with the resolution of no less than 300 dpi [6, 8, 9, 10, 11]. Although some authors [9] examine the possibility of image compression with different JPEG quality factors that mimics input data quality variability, it is not sufficient to imitate smartphone shooting conditions [12].

### 2.2 Method of optical font recognition

Fonts are usually recognized by character sequences (strings, words) without recognizing the characters themselves. Some authors suggest a method of font recognition from a set of more than 2000 classes of fonts using the nearest class mean classifier based on the local feature embedding [13]. Other studies demonstrate the efficient usage of multilayer CNN decomposition and SCAE-based domain adaptation [14]. In these works [13, 14], the training data for the classifier is generated with known public fonts. In numerous studies focused on optical font recognition, the authors analyze exclusively high-resolution images, for example, with a minimal resolution of 400 dpi [15]. In studies [6, 8], appliance of optical font recognition for forgery detection is suggested. Yet it is emphasized by the authors that noise in the images leads to "rapid decrease of recall" and that their method faces difficulty "to accurately classify each character". However, any method for detecting forgery of identity documents should efficiently apply for individual suspicious symbol recognition as the negative impact of missed fakes is tremendous compared to the inconveniences of false alarms. Besides in both [6, 8] a predefined set of typefaces is considered, while our method should classify fonts not represented in the training set.

## 3. PROBLEM STATEMENT AND ITS SPECIAL TRAITS

Let $F$ be the set of all existing fonts. Let $F'$ be the font, specified in standards (for example, OCR-B for machine-readable zones [7]). Let $X = \{I\}$ be the set of character images with the following functions defined on it: $f: X \to F$ – the function of font and the correspondence function $v: X \to \{0,1\}$:

$$v(I) = \begin{cases} 0, f(I) \in F' \\ 1, f(I) \in F \backslash F' \end{cases}$$

We need to develop a classifier, implementing the correspondence function $v$. As the whole set $F$ is too large or even infinite, we should use $F''$, i.e. the set of fonts that are available to create a sample of forged symbols, as an approximation of $F \setminus F'$. Still, the resulting classifier should work properly on any font from $F$ and not only on the subset $F' \cup F''$.

The images of the Russian national passport [16] used as recognition objects have the following special traits:

- Low quality, compression artifacts and low resolution - 96 dpi - as they are captured with smartphone cameras in uncontrolled conditions;

- Complex background and varying quality of paper and printing devices.

## 4. EXPERIMENTS

In our experiments, we use images of digits, i.e. if we consider the recognition alphabet as $A = \{T_i\}_{i=1}^{M}$, in which $M$ - alphabet size, thus in our study $M = 10$. The multi-task learning was applied to train a classifier implementing the function $C: X \to A \times \{0,1\}$ ($C$-type classifier), as it allows to differentiate images by both fonts (i.e. implementing the function $v$ from Sec. 3) and characters. We assume that some characters display more variability between the fonts than the other ones. For instance, the images of '4' in Fig.1 (a) and in Fig.1 (b) differ, while the images of '0' are almost similar. The total number of classes is equal to 20 as $M=10$. To evaluate the acquired classifier, we train its analogue implementing function $C': X \to \{0,1\}$ ($C'$-type classifier) which distinguish only the fonts, so its number of classes is equal to 2.

Similarly to published approaches[6, 8, 9], we consider a sample of images with the font specified in standards (not forged) a negative sample, and a sample with other fonts (forged) - a positive one.

We train CNNs for MRZ and for the Russian national passport ID number, as fonts of these fields are defined in the standards [17, 18], while the other fields have no strict limitations [19]. Although the font for MRZ (called OCR-B) that is used in the Russian national passport is publically available, the Russian national passport is a heavily protected document [20] and is declared a level "A" fraud-proof printing product [21]. It means that everything necessary for its production is classified, including the font used for printing ID numbers [21, 22]. Examples of digits found in the forged Russian national passports are shown in Fig.1 (c), while their analogues printed with the genuine font are demonstrated in Fig.1 (d).

### 4.1 Evaluation of results

False positive results occurring when the genuine passport is recognized as a fake one are less harmful to most financial or legal procedures, such as loan granting procedures or border control, than a situation, in which a fake ID was mistaken for the genuine one.

To evaluate each trained classifier, we calculate its overall sensitivity and specificity on the three test sets (Section 4.4) and build its confusion matrix. However, the entire confusion matrix for a $C$-type classifier is rather large (400 elements in case of 20 classes). Due to the detrimental impact of false negative errors, it seems reasonable to focus only on the part of the confusion matrix corresponding to the forged fonts. For each $C$-type classifier, we build a modification of the confusion matrix where for each class printed with a forged font we provide:

- Number of correctly recognized examples (both the font and the character are recognized correctly, i.e. the forged font is classified as being forged and the character class is determined correctly);

- Number of examples with a correctly recognized font and an incorrectly recognized character;

- Number of examples with an incorrectly recognized font and a correctly recognized character;

- Number of incorrectly recognized examples (both the font and the character are recognized incorrectly).

### 4.2 Network architecture

We suggest using the architecture presented in Fig.2 for $C$-type classifiers. The $C'$-type classifier has the similar architecture except for the fact that the number of neurons in the last fully connected layer equals 2. This architecture has a rather small number of learnable parameters, about 7500 for a $C$-type network with $M=10$, which makes it usable for smartphone applications. The size of the input image is 15x19 pixels.

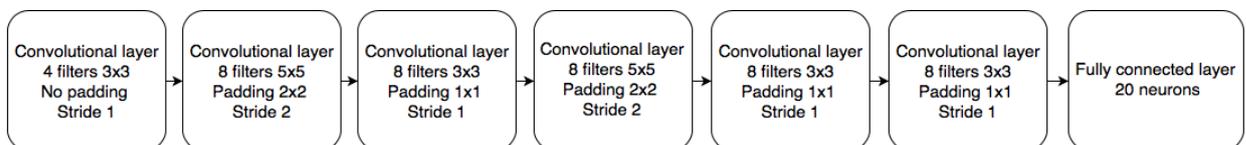

Figure 2. Artificial neural network architecture used for the experiments

### 4.3 Training datasets

In this study, we conduct experiments using digits from two fields of the Russian national passport, applying different approaches to create training datasets. The training set for the classifier for MRZ is synthesized with the method described previously [19]. To create the datasets, we use the fonts available on GoogleFonts [23]. We use OCR-B specified in the MRZ government standards [17, 18] to design a negative sample and 17 fonts similar to OCR-B to generate a positive one. Since the font used in the passport ID number cannot be found in public access, it cannot be used for synthetic data generation. Given this, we use augmented natural data as a negative sample. The positive sample was generated from 46 fonts selected on GoogleFonts [23]. The amount of characters in the training datasets is shown in Fig.3.

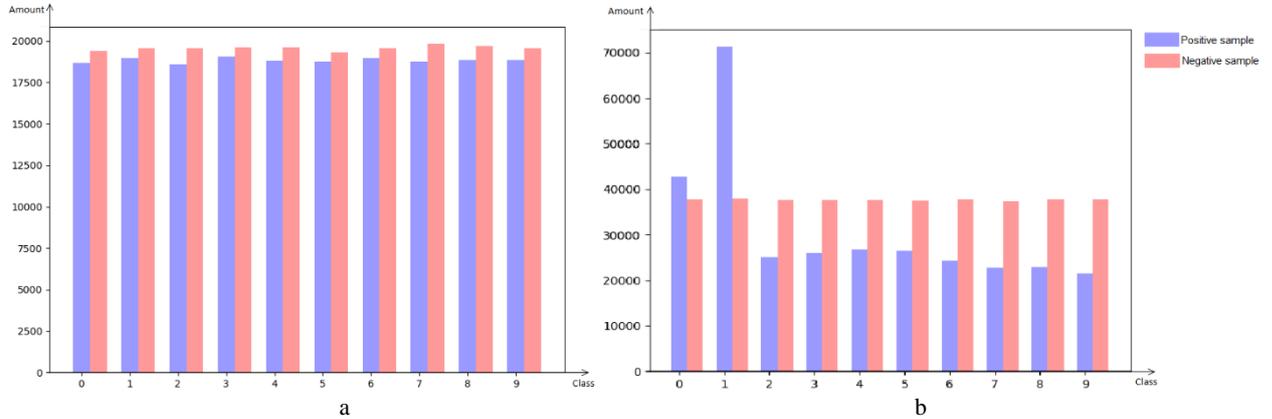

Figure 3. Amount of characters in the training sets for the CNN for MRZ (a) and for the passport ID number (b)

### 4.4 Test datasets

To estimate the resulting CNNs, we use a dataset consisting of 3238 various passport images which were not used in the training process. All images were captured with small-scale digital cameras and were manually labeled. In our experiment, we create three distinct image test sets containing individual characters arranged into 3 subtypes:

- Digits from the passport ID number (129565 images). This sample is used as the negative one to test the CNN for the passport ID number and as the positive one to test the CNN for MRZ;

- Digits from MRZ (68143 images). This sample is used as the negative one to test the CNN for MRZ and as the positive one to test the CNN for passport ID number;

- Digits from dates and authority code (109742 images). This sample contains digits acquired from other passport fields (dates of birth, issue dates, authority codes) and is used as the positive one to test CNNs for both MRZ and the ID number.

It should be emphasized that the fonts used to print dates and authority codes are absent in any training set. Moreover, we do not include the MRZ font in the training set of the ID number classifier and vice versa. The amount of characters in the three test sets is demonstrated in Fig.4.

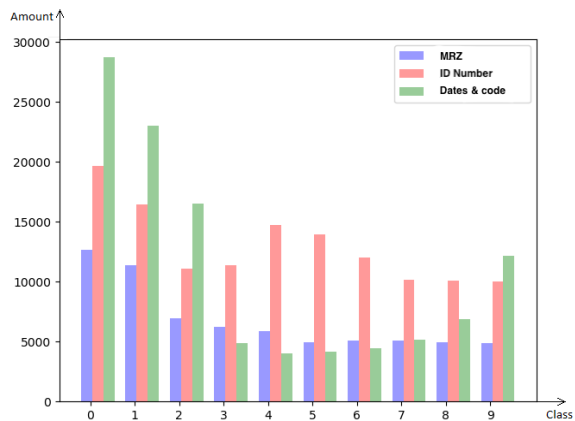

Figure 4. Amount of characters in the test sets

### 4.5 Results of the experiments

#### 4.5.1 Classifiers for passport ID number

First, we train two classifiers for the passport ID number: a $C$-type classifier and a $C'$-type classifier. Table 1 provides the results demonstrated by these CNNs.

In the study, we do not take into account errors of character classification for the $C$-type classifie, i.e. an answer is considered to be True Positive if it belongs to type 1 or 2 from the list in Sec. 4.3. Obviously, the $C$-type classifier surpasses the $C'$-type one in both sensitivity and specificity (Table 1).

Table 1. Results of C-type and $C'$-type classifiers for the passport ID number. TP - character printed with the forged font recognized as such, TN - character printed with the genuine font recognized as such, FP - character printed with the genuine font recognized as a forged one, FN - character printed with the forged font recognized as a genuine one

|  | True Positive | True Negative | False Positive | False Negative | Specificity | Sensitivity |
|---|---|---|---|---|---|---|
| **$C$-type classifer** | **161237** | **113145** | **16379** | **16405** | **87.35%** | **90.77%** |
| $C'$-type classifier | 155976 | 109013 | 20511 | 21666 | 84.16% | 87.80% |

Then, we demonstrate the modification of the confusion matrix described in Sec.4.3 for the forged fonts (Table 2). Here, we denote all the classes by '\_w' to emphasize that we consider only the images with forged ("wrong") fonts. Also, all types of the results are labeled according to the list from Sec.4.3. Our results in Table 2 indicate that error rates of font classification for certain characters are above average. Excluding the symbols with the highest rates of font misclassification, we obtain the sensitivity of the classifier equal to 93.08%. Assuming them being forged, we get the sensitivity reaching 95.12%. In addition, the results provided show that the average error rate of character recognition is 0.88% (maximum is 3.44% for the character '7'). Therefore, we conclude that the proposed CNN can be used for detecting forged fonts, especially if combined with a recognition CNN.

Table 2. Modification of the confusion matrix of $C$-type classifier for the passport ID number and the forged fonts

|  |  | Actual class | | | | | | | | | |
|---|---|---|---|---|---|---|---|---|---|---|---|
|  |  | 0_w | 1_w | 2_w | 3_w | 4_w | 5_w | 6_w | 7_w | 8_w | 9_w |
| Result type | 1 | 34265 | 30917 | 22365 | 10700 | 8767 | 8640 | 8915 | 9331 | 10305 | 15946 |
|  | 2 | 173 | 10 | 59 | 160 | 110 | 116 | 94 | 265 | 27 | 72 |
|  | 3 | 5720 | 2222 | 731 | 1522 | 741 | 1165 | 588 | 533 | 1952 | 738 |
|  | 4 | 45 | 36 | 47 | 29 | 26 | 24 | 37 | 74 | 24 | 151 |

### 4.5.2 Classifiers for MRZ

The results of training the $C$-type and the $C'$-type classifiers for MRZ are provided in Table 3. As well as in Table 1, in Table 3 we do not consider errors of character classification. Unlike the experiment with the ID numbers, the $C'$-type network demonstrates higher sensitivity than the $C$-type one for the forged fonts in MRZ. Yet, it comes with a significant decrease in specificity, as the $C'$-type network classifies more than a half of genuine symbols as forged. The $C$-type classifier demonstrates much higher specificity, however, losing 2.09% (86.28% against 88.37%) in sensitivity.

Table 3. Results of C-type and $C'$-type classifiers for MRZ. TP - character printed with a forged font recognized as such, TN - character printed with the genuine font recognized as such, FP - character printed with the genuine font recognized as a forged one, FN - character printed with the forged font recognized as a genuine one

|  | True Positive | True Negative | False Positive | False Negative | Specificity | Sensitivity |
|---|---|---|---|---|---|---|
| **$C$-type classifer** | **206485** | **53529** | **14614** | **32822** | **78.55%** | **86.28%** |
| $C'$-type classifier | 211477 | 31804 | 36339 | 27830 | 46.67% | 88.37% |

To analyze the errors of $C$-type classifier, we compile Table 4 which is similar to Table 2. Similarly to the measurements for the passport ID number, the results in Table 4 point out that some of the characters in the specified font do not have enough distinguishable features and resemble symbols from the positive sample, especially the symbol '0'. Moreover, if we do not take this character into consideration, the sensitivity reaches 92.56% which outperforms the $C'$-type classifier. If we always consider '0' being forged, the sensitivity is 94.06%. Following these results, we consider the acquired classifier appropriate for the verification of fonts.

Table 4. Modification of the confusion matrix of the $C$-type classifier for MRZ and the forged fonts

|  |  | Actual class | | | | | | | | | |
|---|---|---|---|---|---|---|---|---|---|---|---|
|  |  | 0_w | 1_w | 2_w | 3_w | 4_w | 5_w | 6_w | 7_w | 8_w | 9_w |
| Result type | 1 | 29474 | 37432 | 24025 | 13333 | 13683 | 17581 | 14898 | 14003 | 16523 | 21096 |
|  | 2 | 125 | 646 | 1241 | 320 | 380 | 255 | 861 | 210 | 183 | 216 |
|  | 3 | 18475 | 1216 | 2101 | 2361 | 4432 | 185 | 305 | 846 | 195 | 601 |
|  | 4 | 127 | 188 | 264 | 250 | 229 | 112 | 381 | 248 | 82 | 224 |

## 5. CONCLUSION AND FURTHER WORK

In this paper, we have proposed a method based on optical font recognition to analyze the consistency between the font used in a document and the one specified in standards. We consider the cases, in which all the fonts necessary for classifier training are publically available and a sample of the font specified in standards is obtained from images of real documents. The test sets consist of images of digits from different fields of Russian national passports. All test images were captured with small-scale digital cameras. The most important trait of the suggested classifier is its generalization ability, confirmed by the fact that the fonts from the test sets used as forged ones were not present in the training set. We use multi-task learning to train a CNN (called a $C$-type classifier) that recognizes both the character class and the authenticity of the font. Also, we train another classifier that recognizes only the font (called $C'$-type classifier) to compare these two approaches. The obtained results indicate that:

- The classifiers that are distinguishing classes by both characters and fonts ($C$-type) demonstrate higher sensitivity and specificity than the ones trained only to predict the font class ($C'$-type);

- The suggested $C$-type classifiers can be used for detection of the forged fonts, but different characters should be analyzed with different weights as some of them do not have enough features. We propose that, when embedded into a document recognition system, the suggested classifiers should be rather used as a part of an authenticity control system than individually.

We expect that the combination of a standard character classifier and the suggested classifier should be efficient on character sequences (e.g. the Russian passport ID number consists of 10 digits). For instance, such a combination can operate as follows:

1. Recognize all symbols with the standard classifier and set weights according to the recognition results;
2. Recognize each symbol with the suggested classifier and, if an unsuitable font is detected or the result of symbol recognition differs from the result of step 1, consider the symbol to be printed with an inadequate font;
3. Evaluate the symbol sequence according to the results of step 2 and the weights from step 1 to make a verdict about the field authenticity.

For future work, we plan to study the applicability of the suggested method for different identity documents with specific fonts, such as French ID cards or British driving licenses. Besides, we will explore the possibility of training $C'$-type classifiers to a sufficient sensitivity and specificity. This type of classifiers does not have limitations on the number of character classes as the number of neurons in its last layer is always equal to 2, unlike the $C$-type.

## AKNOWLEDGMENTS

The reported study was partially funded by RFBR according to the research projects 17-29-03236, 17-29-07093.

## REFERENCES

[1] Y. bin Kwon and J. hoon Kim, "Recognition based verification for the machine readable travel documents", in International Workshop on Graphics Recognition (GREC 2007).
[2] L. Koker, "Money laundering compliance the challenges of technology", (06 2016).
[3] A. Gerasimov, "Digital fraud: Risks and damages", URL: https://bosfera.ru/bo/cifrovoe-moshennichestvo-riski-i-ushcherb
[4] Equifax and EFX, "The New Reality of Synthetic ID Fraud", (Atlanta, Georgia, 2015).
[5] "A survey on fintech", J. Netw. Comput. Appl. 103, 262-273, (Feb. 2018).

[6] R. Bertrand, O.R. Terrades, P. Gomez-Krämer, P. Franco and J.-M. Ogier, "A conditional random field model for font forgery detection", 2015 13th International Conference on Document Analysis and Recognition (ICDAR), 576-580 (2015).

[7] "ICAO Doc. 9303", URL: https://www.icao.int/publications/pages/publication.aspx?docnum=9303

[8] R. Bertrand, P. Gomez-Krämer, O.R. Terrades, P. Franco and J.-M. Ogier, "A system based on intrinsic features for fraudulent document detection", 2013 12th International Conference on Document Analysis and Recognition (ICDAR), 106-110 (2013).

[9] S. Abramova and R. Böhme, "Detecting copy-move forgeries in scanned text documents", (2016).

[10] A. Zramdini and R. Ingold "Optical font recogntion using typographical features", IEEE Trans. Pattern Anal. Match. Intell. 20, 877-882 (Aug. 1998).

[11] A. Satkhozhina, I. Ahmadullin, and J.P. Allebach, "Optical font recognition using conditional random field", in Proceedings of the 2013 ACM Symposium on Document Engineering, DocEng'13, 119-122, ACM, New York, NY, USA (2013). DOI: 10.1145/2494266.2494307.

[12] K. Bulatov, V. Arlazarov, T. Chernov,et al., "SmartIDReader: Document recognition in video stream," in 14th IAPR International Conference on Document Analysis and Recognition (ICDAR), vol. 6, IEEE, 2017,pp. 39–44. DOI: 10.1109/ICDAR.2017.347.

[13] G. Chen, J. Yang, H. Jin, J. Brandt, E. Shechtman, A. Agarwala, and T.X. Han, "Large-scale visual font recognition", in Proc. IEEE Conference on Computer Vision and Pattern Recognition (CVPR), (2014).

[14] Z. Wang, J. Yang, H. Jin, E. Shechtman, A. Agarwala, J. Brandt, and T.S. Huang, "Deepfont: Identify your font from an image", CoRR abs/1507.03196 (2015).

[15] A. Berenguel, O.R. Terradesm J. Lladós, and C. Cañero, "E-counterfeit: a mobile-server platform for document counterfeit detection", CoRR abs/1708.06126 (2017).

[16] Government of the Russian Federation, "Order N 828 On the approval of the provision of a passport of a citizen of the Russian Federation", (July 8, 1997).

[17] Ministry of Internal Affairs of the Russian Federation, "Order N 851 On the approval of administrative regulations of the ministry of internal affairs of the Russian Federation for the provision of state service for issuing, replacing of the passports of a citizen of the Russian Federation, which certify the identity of the citizen of the Russian Federation on the territory of the Russian Federation" (November 13, 2017).

[18] Government of the Russian Federation, "Order N 424 On the machine readable zone in a passport of a citizen of the Russian Federation", (May 27, 2011).

[19] Y. Chernyshova, A. Gayer, and A. Sheshkus, "Generation method of synthetic training data for mobile OCR system", Proc. SPIE 10696, 10696-10696-7 (2018). DOI: 10.1117/12.2310119.

[20] T. Chernov, S. Kolmakov, and D. Nikolaev, "An algorithm for detection of phase estimation of protective elements periodic lattice on document image", Pattern Recognition and Image Analysis, 53-65 (2017). DOI: 10.1134/S1054661817010023.

[21] Ministry of Finance of the Russian Federation, "Order N 14 On the implementation of the Government Decree on November 11, 2002 N 817" (February 7, 2003).

[22] Government of the Russian Federation, "Decree N 727 On licensing of activity of manufacture and realization of the polygraphic production protected from fakes" (July 29, 2016).

[23] "Google fonts", URL: https://fonts.google.com.